\documentclass[lettersize,journal]{IEEEtran}
\IEEEoverridecommandlockouts
\usepackage{cite}
\usepackage{amsmath,amssymb,amsfonts,siunitx}
\usepackage{algorithmic}
\usepackage{graphicx}
\usepackage{multirow}  
\usepackage{textcomp}
\usepackage{xcolor}
\usepackage{color}
\usepackage{float}
\usepackage{booktabs}
\usepackage{subfigure}
\usepackage{subfloat}
\usepackage{enumitem}
\usepackage[colorlinks, citecolor=blue]{hyperref}
\usepackage[misc]{ifsym}
\usepackage{eso-pic}
\begin{document}
\AddToShipoutPictureFG*{ 
  \AtPageUpperLeft{ 
    \hspace{0.15\textwidth} 
    \raisebox{-1.5cm}{ 
      \parbox{0.88\textwidth}{ 
        \footnotesize \textit{This work has been submitted to the IEEE for possible publication. Copyright may be transferred without notice, after which this version may no longer be accessible.}
      }
    }
  }
}
\title{Recovering Partially Corrupted Objects via Sketch-Guided Bidirectional Feature Interaction}

\author{Yongle Zhang, Yimin Liu, Yan Huang, Qiang Wu,~\IEEEmembership{Senior Member,~IEEE}

\thanks{Yongle Zhang and Qiang Wu are with the School of Electrical and Data Engineering, and Yan Huang is with the Australian Artificial Intelligence Institute, University of Technology Sydney, NSW 2007, Australia (e-mail: yongle.zhang@student.uts.edu.au; \{yan.huang-7, qiang.wu\}@uts.edu.au).}
\thanks{Yimin Liu is with the School of Computer Science and Information Engineering, Hefei University of Technology, Hefei 230009, China (e-mail: yiminliu@mail.hfut.edu.cn). (Corresponding author: Yimin Liu)}
}
\maketitle
\begin{abstract}

Text-guided diffusion models have achieved remarkable success in object inpainting by providing high-level semantic guidance through text prompts. However, they often lack precise pixel-level spatial control, especially in scenarios involving partially corrupted objects where critical uncorrupted cues remain. To overcome this limitation, sketch-guided methods have been introduced, using either indirect gradient modulation or direct sketch injection to improve structural control. Yet, existing approaches typically establish a one-way mapping from the sketch to the masked regions only, neglecting the contextual information from unmasked object areas. This leads to a disconnection between the sketch and the uncorrupted content, thereby causing sketch-guided inconsistency and structural mismatch. To tackle this challenge, we propose a sketch-guided bidirectional feature interaction framework built upon a pretrained Stable Diffusion model. Our bidirectional interaction features two complementary directions, context-to-sketch and sketch-to-inpainting, that enable fine-grained spatial control for partially corrupted object inpainting. In the context-to-sketch direction, multi-scale latents from uncorrupted object regions are propagated to the sketch branch to generate a visual mask that adapts the sketch features to the visible context and denoising progress. In the sketch-to-inpainting direction, a sketch-conditional affine transformation modulates the influence of sketch guidance based on the learned visual mask, ensuring consistency with uncorrupted object content. This interaction is applied at multiple scales within the encoder of the diffusion U-Net, enabling the model to restore object structures with enhanced spatial fidelity. Extensive experiments on two newly constructed benchmark datasets demonstrate that our method achieves superior sketch-guided consistency compared to state-of-the-art approaches.

\end{abstract}
\section{Introduction}
\label{sec:introduction}
\begin{figure}[t]
\centerline{\includegraphics[width=0.48\textwidth]{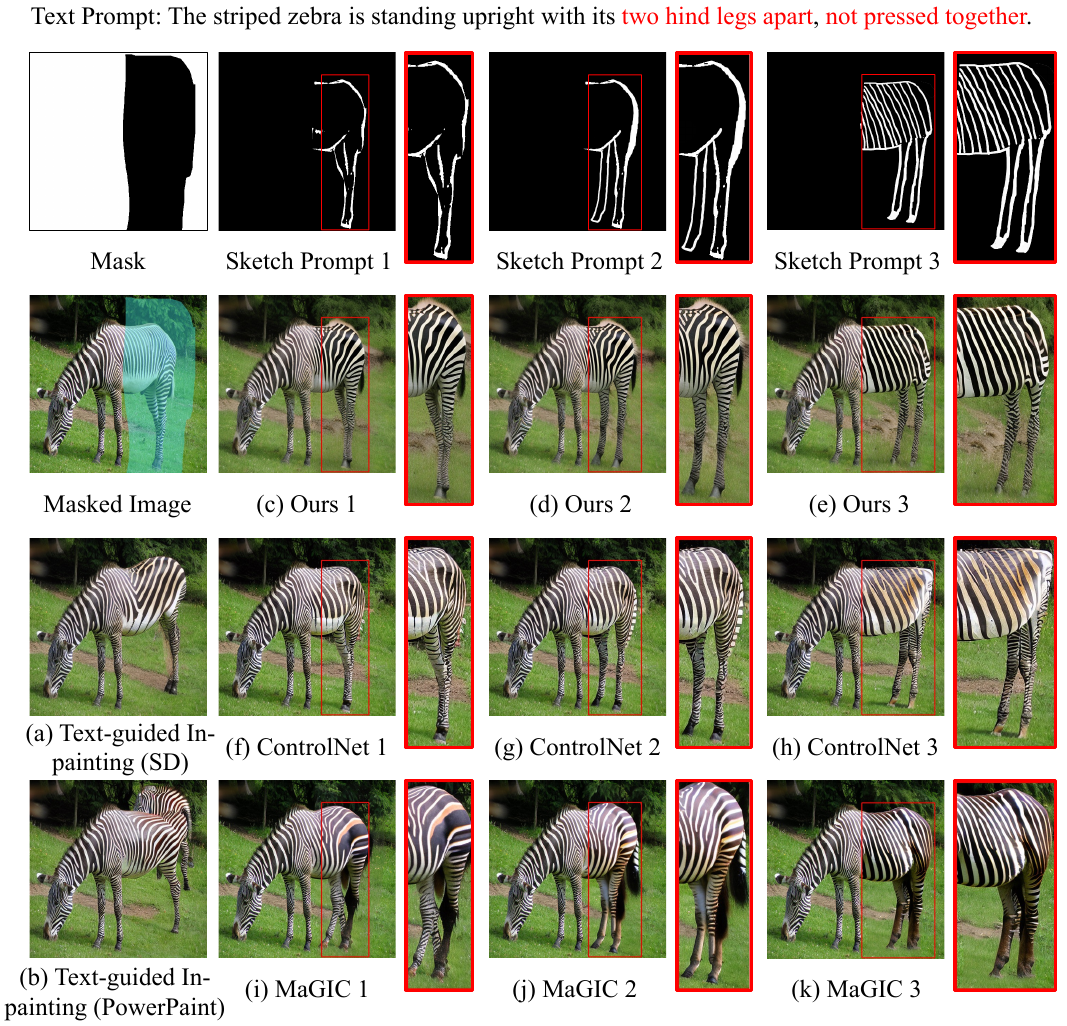}}
\setlength{\abovecaptionskip}{-1pt}
\caption{Inpainted results from different diffusion-based methods: (a,b) Text-guided results generated by Stable Diffusion~\cite{rombach2022high} and PowerPaint~\cite{zhuang2025task}, which struggle to produce plausible object structures within the masked regions; (c–e) Results from our sketch-guided method using corresponding sketch prompts, demonstrating superior structural accuracy and sketch guidance consistency compared to ControlNet and MaGIC; (f–h) Results from ControlNet guided by sketch prompts~\cite{zhang2023adding}; (i–k) Results from MaGIC guided by sketch prompts~\cite{wang2024magic}. [Best view with zoom-in.]}
\vspace{-0.3cm}
\label{Figure:threetypesofinpainting}
\end{figure}

\IEEEPARstart{I}{mage} completion, or inpainting~\cite{10.1145/344779.344972, sun2005image}, aims to restore missing regions in an image with content that is both visually plausible and semantically coherent. This technique underpins a wide range of applications such as object recovery and image editing~\cite{criminisi2004region, ruvzic2014context, huang2025diffusion}. The task becomes particularly challenging when dealing with partially corrupted objects, where substantial portions of an object are missing while other parts remain intact (e.g., the masked image shown in Fig.~\ref{Figure:threetypesofinpainting}). In such cases, critical semantic details may be obscured, both from the object itself and its surrounding context. However, the visible, uncorrupted regions often contain rich visual cues that, if properly utilized, can aid in accurate reconstruction. However, existing diffusion-based inpainting methods often fail to fully exploit these cues.

\begin{figure}[t]
\centerline{\includegraphics[width=0.48\textwidth]{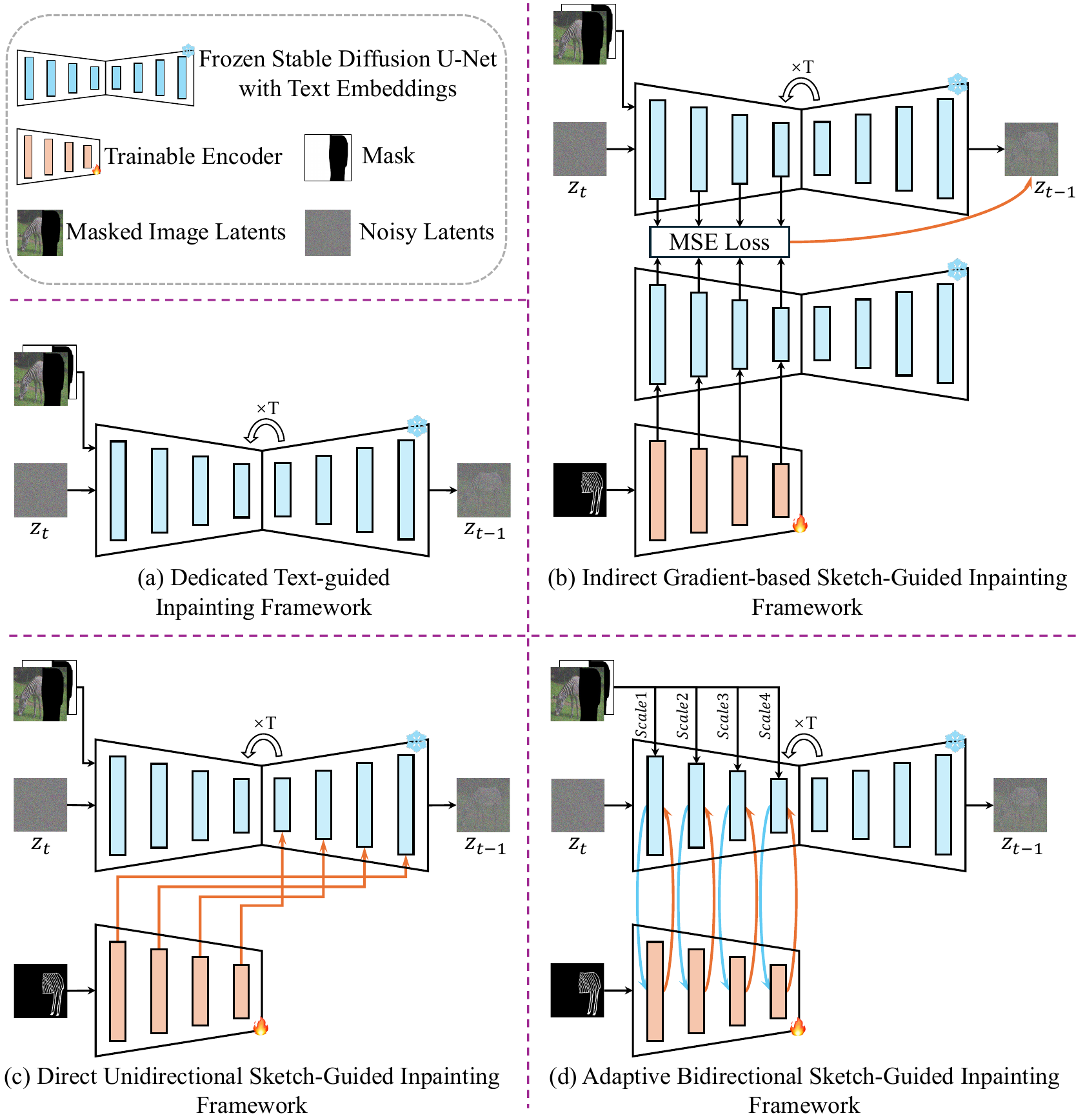}}
\setlength{\abovecaptionskip}{-1pt}
\caption{Comparison of diffusion-based object inpainting frameworks: (a) text-guided~\cite{rombach2022high,zhuang2025task, xie2023smartbrush}; (b) sketch-guided with indirect gradient-based guidance~\cite{wang2024magic}; (c) sketch-guided with direct unidirectional guidance~\cite{zhang2023adding}; (d) our adaptive bidirectional sketch-guided approach.}
\vspace{-0.3cm}
\label{Figure:threetypesofinpaintingframework}
\end{figure}

Recent advances in diffusion-based text-to-image models~\cite{ho2020denoising, li2022blip, li2023blip, ramesh2022hierarchical, saharia2022photorealistic, rombach2022high} have demonstrated remarkable capabilities in generating high-quality images conditioned on text prompts. These developments have also led to significant progress in text-guided diffusion pipelines for object inpainting. Typically, such dedicated pipelines extend the input channels of existing diffusion models to incorporate both the corrupted image and its corresponding mask, thereby enabling text-guided object inpainting~\cite{saharia2022palette, wang2023imagen, rombach2022high, ju2024brushnet, manukyan2023hd, xie2023smartbrush, zhuang2025task, yang2023uni}, as illustrated in Fig.~\ref{Figure:threetypesofinpaintingframework} (a). While these approaches are effective in synthesizing semantically plausible content within the masked regions based on the provided text, they often lack fine-grained, pixel-level control over object structures or spatial postures, even when the text prompts are highly detailed. For instance, in Fig.~\ref{Figure:threetypesofinpainting} (a), although the text explicitly specifies ``two hind legs apart, not pressed together,'' the reconstructed zebra still exhibits closely positioned hind legs. This illustrates a critical limitation: while text prompts excel at conveying high-level semantics, their abstract concepts often struggle to translate into spatially pixel-wise reconstructions such as precise postures or orientations. To address this, recent shape-aware inpainting methods~\cite{xie2023smartbrush, zhuang2025task, chen2025improving, chiu2024brush2prompt} have emerged, incorporating object-specific masks and label prompts to steer diffusion-based object inpainting. However, when the uncorrupted regions of an object exhibit distinctive structural characteristics such as the specific posture of an animal, these approaches often fail to recover object details that align well with the visible, uncorrupted object regions (see Fig.~\ref{Figure:threetypesofinpainting} (b)). 

As the saying goes, ``a picture is worth a thousand words.'' Motivated by this, several recent studies have explored the use of visual prompts such as sketches~\cite{xing2023diffsketcher,vinker2023clipascene} to offer fine-grained spatial control in text-guided diffusion inpainting. These sketches serve as explicit spatial cues that convey the object's structure or orientation and are incorporated into the diffusion process to guide the reconstruction more precisely. Existing sketch-guided inpainting methods can be broadly categorized into two groups: indirect gradient-based guidance~\cite{wang2024magic, voynov2023sketch} and direct feature integration~\cite{zhang2023adding, kim2023reference}.

Gradient-based methods guide the denoising process by computing gradients between the model's latent features and a target sketch map. As shown in Fig.~\ref{Figure:threetypesofinpaintingframework} (b), a trainable encoder extracts features from the input sketch, which are then compared to the inpainting latents using an MSE loss. The resulting gradients are back-propagated to adjust the denoised latents, steering them toward alignment with the sketch. While this strategy enables the model to follow the sketch to some extent, its indirect nature often leads to unstable training and suboptimal alignment with the sketch guidance (see Fig.~\ref{Figure:threetypesofinpainting} (i)$\sim$(k)). Moreover, when the object is only partially corrupted, these methods typically restrict the sketch guidance to the masked region, overlooking the correlations between the sketch and the unmasked portions of the object. This limitation often results in inconsistent or fragmented reconstructions.

Direct integration methods mitigate the limitations of gradient-based approaches by injecting sketch features directly into the inpainting model, typically through operations such as element-wise addition. For instance, ControlNet~\cite{zhang2023adding}, shown in Fig.~\ref{Figure:threetypesofinpaintingframework} (c), employs a trainable encoder to extract sketch embeddings, which are then incorporated into a pre-trained text-guided diffusion model. While this approach circumvents the instability associated with gradient-based guidance, it still encounters challenges in scenarios involving partial corruption. Specifically, ControlNet unilaterally integrates partial sketch features without adapting them to the uncorrupted object regions during the denoising inpainting process. As a result, the guidance remains ambiguous and often inconsistent with the visible object context. (see Fig.~\ref{Figure:threetypesofinpainting} (f)$\sim$(h)).

To overcome the aforementioned challenges, we propose an \textbf{adaptive bi-directional sketch-guided inpainting framework} specifically tailored for partially corrupted object inpainting. Our approach builds upon a frozen, pre-trained text-guided Stable Diffusion model, but introduces three innovations over direct-integration methods such as ControlNet (see Fig.~\ref{Figure:threetypesofinpaintingframework} (d)): (1) \textbf{Bi-directional feature interaction}. Before incorporating sketch features, we first integrate multi-scale latents extracted from the masked image and its corresponding mask into the denoised latents of the diffusion model. These fused latents are then passed to the sketch branch (indicated by the blue arrows in Fig.~\ref{Figure:threetypesofinpaintingframework} (d)), allowing the sketch representation to dynamically adapt to both the uncorrupted object contexts and the progression of the denoising process. To facilitate this, we propose a context-aware feature fusion module that learns a \textit{visual mask} based on the fused object latents and the guided sketch. (2) \textbf{Sketch feature modulation}. Instead of direct element-wise addition, we introduce a sketch-conditional affine transformation modulated by the learned \textit{visual mask}. This mechanism enables adaptive control over the influence of sketch features, promoting fine-grained guidance while maintaining consistency with the visible parts of the object. (3) \textbf{Early-stage integration}. We embed the bi-directional interaction module into the encoder stage of the diffusion U-Net, enabling early influence over the object's structural fidelity. The guided sketch features are then propagated to the decoder through skip connections, ensuring continuity and alignment throughout the inpainting process.

Lastly, while most existing sketch-guided diffusion inpainting methods overlook the unique challenges posed by partially corrupted object restoration, our work explicitly targets this scenario. To support our research and facilitate future developments in this area, we introduce two novel datasets—CUB-sketch and MSCOCO-sketch—each containing four-tuple annotations (text description, partial mask, partially masked image, and partial sketch) for every image sample. In summary, our contributions are as follows:
\begin{itemize}
\item We propose a novel diffusion-based sketch-guided inpainting framework featuring bi-directional feature interaction, specifically tailored for partially corrupted object restoration. 
\item We introduce a context-aware feature fusion mechanism and a sketch-conditional affine transformation module to realize bi-directional interaction, enabling adaptive integration of sketch information conditioned on uncorrupted object regions. 
\item We release two benchmark datasets, CUB-sketch and MSCOCO-sketch, which provide paired annotations of sketches, masks, and text to support standardized evaluation and future research in this domain.
\end{itemize}

\section{Related Work}
\label{sec: related work}
\subsection{Image Inpainting with Text Prompts}\label{subsec21}
Traditional image inpainting methods~\cite{iizuka2017globally, zuo2018minimum, yu2019free, zhang2023mutual, quan2022image, deng2023context} often utilize CNNs~\cite{krizhevsky2012imagenet} or GANs~\cite{goodfellow2014generative} to reconstruct missing areas by relying on residual information from uncorrupted regions. However, when recovering a partially masked object within an image, these methods often fail to generate reliable semantic content due to limited contextual information. Text-guided image inpainting~\cite{zhang2020text, zhang2020textneural, lin2020mmfl, wu2021adversarial, ni2023nuwa, wang2023imagen, avrahami2023blended} addresses this limitation by using text prompts to supplement missing semantics and guide content restoration. Early efforts~\cite{zhang2020text, zhang2020textneural, lin2020mmfl, wu2021adversarial} focus on the modality gap between text and image features while then integrating textual guidance into missing image spaces. Despite these advances, achieving coherent results that align with text prompts remains challenging.

Recent diffusion models~\cite{ho2020denoising, songdenoising} trained on large-scale image-text datasets have demonstrated excellent generative capabilities in text-to-image (T2I) tasks~\cite{ramesh2022hierarchical, saharia2022photorealistic, rombach2022high}. Many methods utilize the strong priors of these T2I models and fine-tune them for text-guided image inpainting~\cite{saharia2022palette, wang2023imagen, avrahami2023blended, rombach2022high, ju2024brushnet, manukyan2023hd}. Specifically, text prompts encoded by CLIP~\cite{radford2021learning} guide semantic content restoration through cross-attention layers in diffusion U-Nets. For example, SD-Inpainting~\cite{rombach2022high} integrates the mask, damaged image, and noisy latent to fine-tune Stable Diffusion for text-guided object inpainting. Similarly, BrushNet~\cite{ju2024brushnet} incorporates mask information and unmasked context into the hierarchical noisy feature layers of Stable Diffusion~\cite{rombach2022high}, achieving improved text-guided consistency. However, these methods struggle to deliver controllable and fine-grained pixel-level reconstruction of object structures guided solely by textual prompts. 

Shape-aware completion methods like SmartBrush~\cite{xie2023smartbrush} and PowerPaint~\cite{zhuang2025task} focus on filling semantic content within object-shaped missing regions using text prompts. However, these approaches often yield ambiguous semantics for partially occluded object masks, as text prompts alone fail to establish precise pixel-level visual correlations between corrupted and uncorrupted object parts. Our proposed method overcomes this limitation by integrating visual cues from partial sketches into frozen pretrained text-guided diffusion models to guide structural generation in partially corrupted object regions while adapting to surrounding regions.

\subsection{Image Inpainting with Visual Prompts}\label{subsec22} 
Visual prompts have shown effectiveness across various vision tasks~\cite{wang2024sam,yuan2025open,zhang2023adding,mou2024t2i,jiang2022text2human,zuo2025learning}. For example, SAM-CLIP~\cite{wang2024sam} and Open-Vocabulary SAM~\cite{yuan2025open} combine SAM's visual bounding box~\cite{kirillov2023segment} with CLIP text prompts~\cite{radford2021learning} for segmentation and recognition tasks. In T2I tasks, models like ControlNet~\cite{zhang2023adding}, T2I-Adapter~\cite{mou2024t2i}, and Text2human~\cite{jiang2022text2human} use visual inputs (e.g., sketches, semantic maps, depth maps, pose maps) to guide diffusion-based image generation. Unlike image completion, which is constrained by existing uncorrupted object content, these T2I methods excel at synthesizing visually coherent objects entirely from scratch.

Certain existing image inpainting methods \cite{wang2024magic, Sharma_2024_CVPR}, including fine-tuned ControlNet inpainting \cite{zhang2023adding}, leverage structural guidance such as sketches, similar to those used in text-to-image (T2I) frameworks. These T2I-inspired methods perform effectively when the object is entirely missing from the corrupted region. However, they struggle when the object is only partially corrupted, due to a lack of perceptual understanding of the visible, uncorrupted object context and insufficient integration of structural information. In contrast, our proposed method—featuring sketch-guided bi-directional feature interaction—overcomes this limitation. This bi-directional mechanism not only enables fine-grained sketch guidance for inpainting partially corrupted objects but also adapts to the uncorrupted object context during the guidance process. Consequently, our approach facilitates precise object structure control and ensures sketch-guided coherence.

\begin{figure*}[h]
\centering
\includegraphics[width=0.805\textwidth]{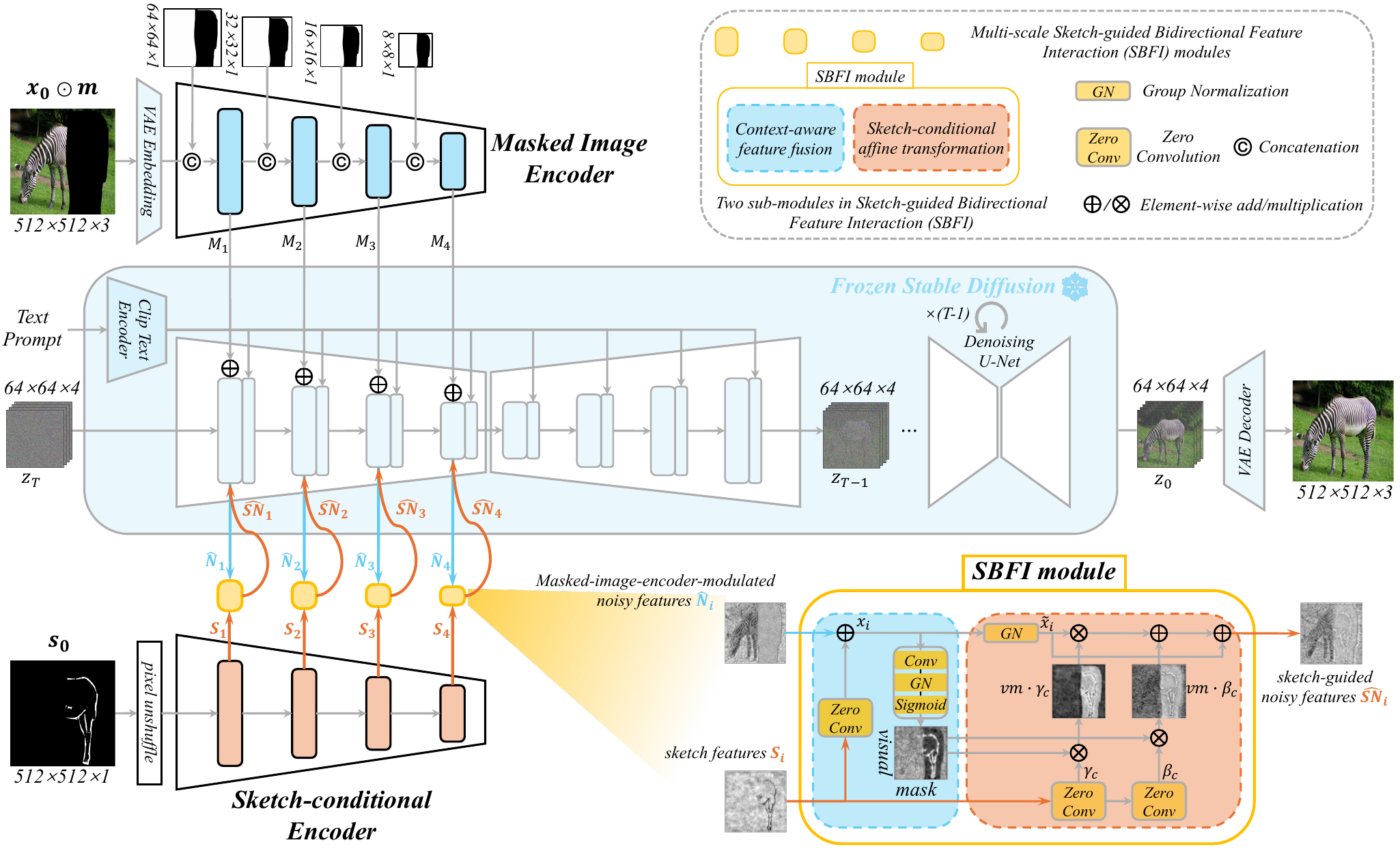} 
\setlength{\abovecaptionskip}{-1pt}
\caption{Our pipeline builds on the frozen T2I Stable Diffusion model and incorporates the following key components: a masked image encoder, which integrates binary mask localization and object contextual information from the corrupted image into the noisy features; and a sketch-conditional encoder followed by multi-scale Sketch-guided Bidirectional Feature Interaction (SBFI) modules, which enable fine-grained sketch integration while adapting to the uncorrupted object context during the integration process.}
\vspace{-0.2cm}
\label{fig:fullnetwork}
\end{figure*}
\section{Methodology}
\label{sec:methodology}

As shown in Fig.~\ref{fig:fullnetwork}, to enable partially corrupted object inpainting using partial sketch guidance within a frozen text-embedded Stable Diffusion model, we introduce a \textit{Masked Image Encoder} that extracts multiscale features from the corrupted image. These features encode both the spatial localization of the mask and contextual cues from the visible, uncorrupted regions. The resulting representations are then injected into the denoising latents of the Stable Diffusion model, yielding multiscale noisy features modulated by the masked image encoder.

In parallel, a \textit{Sketch-Conditional Encoder} extracts multiscale features from the partial sketch. At the core of our framework is the \textit{Sketch-guided Bidirectional Feature Interaction (SBFI)} module, which first fuses sketch-derived features and masked-image-encoder-modulated features at each encoder level via a \textit{context-aware feature fusion sub-module}. This fusion enables the sketch guidance to dynamically adapt to both the uncorrupted object context and the current state of the denoising process. Subsequently, a \textit{sketch-conditional affine transformation sub-module} modulates the influence of the sketch features based on a predicted visual mask obtained from the fusion stage. The resulting sketch-guided noisy features are then reintegrated into the frozen Stable Diffusion model.

The SBFI module operates at multiple scales within the encoder stage of the diffusion model, enabling fine-grained structural control and sketch-guided consistency from the early stages of generation. Meanwhile, the embedded text prompt—leveraged by the frozen diffusion model—ensures that the reconstructed object remains semantically aligned with the specified description. The detailed components of the framework are described in the following subsections.

\subsection{Preliminaries}\label{subsec31}
Our pipeline builds on the pre-trained text-to-image diffusion model, Stable Diffusion \cite{rombach2022high}, which comprises a variational autoencoder (VAE) and a UNet denoiser. The VAE encodes a clean image $x_{0}$ into a latent space $z_{0}$ and decodes it for image reconstruction. The UNet denoiser conducts diffusion in the latent space through a forward and reverse process.

In the forward process, Gaussian noise $\epsilon$ is added to the clean latent image $z_{0}$ to generate a noisy sample $z_{t}$ at timestep $t$ as follows:
\begin{equation}
  z_{t}=\sqrt{\bar{\alpha _{t} } }z_{0}+  \sqrt{1-\bar{\alpha _{t} }  } \epsilon, \quad \epsilon \sim \mathcal{N}(0, I), \label{eq:forwardprocess}
\end{equation}
where $\bar{\alpha _{t}}$ denotes the noise level. In the reverse process, the learnable UNet denoiser $\epsilon_{\theta}$ predicts the added noise $\epsilon_{t}$ at each timestep $t$, conditioned on the input text, enabling step-by-step denoising from Gaussian noise. The training objective for the diffusion model is defined as:
\begin{equation}  \mathcal{L}=\mathbb{E}_{z_{0},t,\epsilon_{t}}\left\|\epsilon_{t}-\epsilon_{\theta}\left(z_{t}, \tau_{\theta}(text), t\right)\right\|_{2}^{2}, \label{eq:obj_sd}
\end{equation}
where $\tau_{\theta}$ is the CLIP text encoder. Building on this pre-trained model, our fine-tuning objective incorporates the trainable masked image encoder $enc_{\theta}(m)$ and sketch-conditional encoder $enc_{\theta}(s)$, resulting in the following loss: 
\begin{equation}  \mathcal{L}=\mathbb{E}_{z_{0},t,\epsilon_{t},m,s}\left\|\epsilon_{t}-\epsilon_{\theta}\left(z_{t}, \tau_{\theta}(text),enc_{\theta}(m), enc_{\theta}(s),t\right)\right\|_{2}^{2}. \label{eq:obj_our}
\end{equation}

\subsection{Masked Image Encoder}\label{subsec32}
To fine-tune Stable Diffusion for sketch-guided image inpainting, we introduce a Masked Image Encoder, as shown in the top-left corner of Fig.~\ref{fig:fullnetwork}. This encoder captures critical information from the masked image, including the details of uncorrupted object regions and the spatial location of the partial mask. These features are propagated to the multi-scale noisy feature maps within the UNet encoder, providing rich contextual cues that facilitate the subsequent integration of partial sketch guidance into the corrupted object area.

The architecture of the masked image encoder mirrors that of the UNet denoiser encoder, except that the text cross-attention layers are removed. The pretrained weights from the original denoiser encoder are retained, serving as strong priors for extracting features from the masked image. Given a $512\times512\times3$ masked image $x_{0}\odot m$, where the binary mask $m$ with pixel value 0 indicates the corrupted regions, we first employ the VAE encoder to embed the image into a $64\times64\times4$ latent space aligned with the latent data distribution of the pre-trained denoiser UNet. 

The masked image encoder then extracts a set of multi-scale features $M = \left \{ M_{1},M_{2},M_{3},M_{4} \right \}$. To reinforce spatial awareness of the corrupted regions, the corresponding scale binary mask is concatenated channel-wise at each feature level.
The dimensions of $M$ are aligned with the intermediate noisy features $N = \left \{ N_{1},N_{2},N_{3},N_{4} \right \}$ within the denoiser's encoder. At each scale, the masked features $M$ are added to the corresponding noisy features $N$, yielding modulated features $\hat{N}_i$. The overall process is defined as:
\begin{align}
M &= enc_{\theta } \left(vae_{emb} (x_{0} \odot m), \downarrow_{8,16,32,64} (m) \right), \label{eq:M} \\
\hat{N}_{i} &= N_{i} + M_{i}, \quad i \in \{1, 2, 3, 4\}, \label{eq:Ni}
\end{align}
where $\hat{N}_{i}$ denotes the masked-image-encoder-modulated noisy features, as illustrated by the blue arrows in Fig.~\ref{fig:fullnetwork}. These modulated features encode both the preserved visual information from uncorrupted regions and the spatial location of the mask.

\begin{figure*}[t]
  \centering
  \setlength{\abovecaptionskip}{-1pt}
  \includegraphics[width=0.75\textwidth]{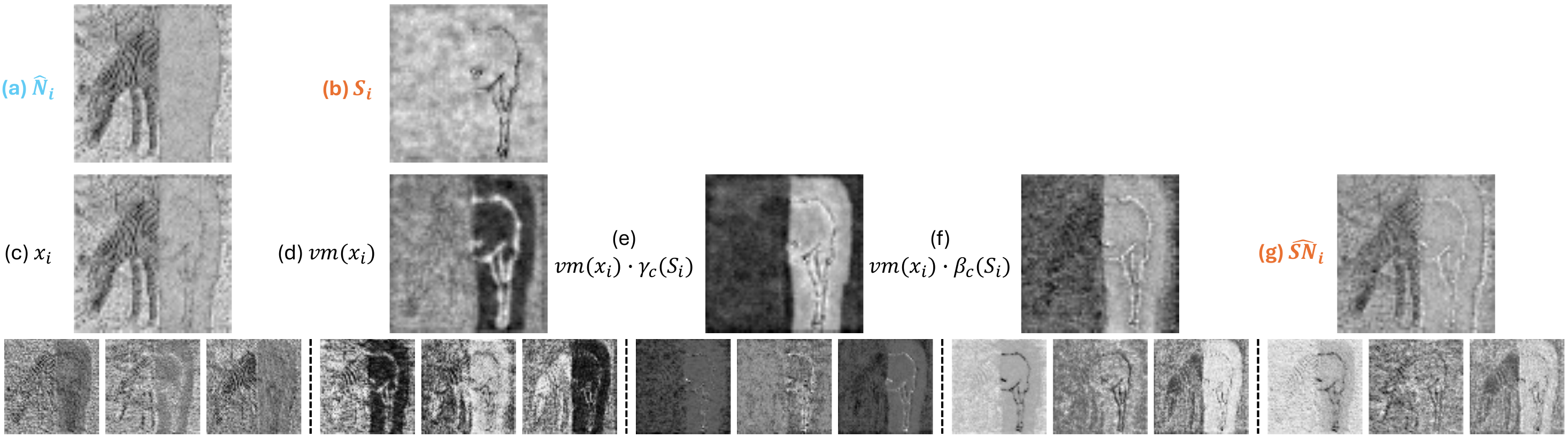}
  \caption{The visualization of features generated by the Sketch-guided Bidirectional Feature Interaction (SBFI) module, including: (a) masked-image-encoder-modulated noisy features $\hat{N}_{i}$, (b) sketch features $S_{i}$, (c, d) features from the context-aware feature fusion sub-module, and (e$\sim$g) features from the sketch-guided affine transformation sub-module. Here, we use the scale factor $i=1$ in the multi-scale SBFI module as an example to visualize the overall feature distribution by averaging all channels, as shown in the first two rows. Additionally, in the third row, we visualize the local distribution patterns by displaying the first three individual channels of the corresponding features in (c$\sim$g).}
  \label{fig:exp_vssavisulize}
\end{figure*}
\subsection{Sketch-Conditional Encoder with Sketch-guided Bidirectional Feature Interaction}\label{subsec33}
We propose a Sketch-Conditional Encoder augmented with multi-scale Sketch-guided Bidirectional Feature Interaction (SBFI) modules to inject partial sketch features into the corrupted regions of the noisy features. The architecture of the sketch-conditional encoder mirrors that of the masked image encoder, as shown in the bottom-left corner of Fig.~\ref{fig:fullnetwork}.

\textbf{\textit{Sketch-Conditional Encoder.}} Given a $512\times512\times1$ partial sketch image $s_{0}$, which contains structural cues for the partially corrupted region, we first downsample it to a $64\times64\times1$ resolution using pixel unshuffle \cite{shi2016real}. These downsampled features are then passed through the sketch-conditional encoder to extract multi-scale sketch features $S=\left \{ S_{1}, S_{2}, S_{3}, S_{4} \right \}$, whose dimensions match those of the masked-image-encoder-modulated noisy features $\hat{N}= \left \{ \hat{N}_{1}, \hat{N}_{2}, \hat{N}_{3}, \hat{N}_{4} \right \}$. The sketch features are fused with the corresponding modulated noisy features at each scale via the proposed SBFI module, denoted by $f_{sbfi}$, as follows:
\begin{align}
S &= enc_{\theta } \left(s_{0} \right), \label{eq:S} \\
\hat{SN}_{i} &= f_{sbfi}( \hat{N}_{i},S_{i}), \quad i \in \{1, 2, 3, 4\}. \label{eq:SN}
\end{align}

\textbf{\textit{Sketch-guided Bidirectional Feature Interaction.}} The SBFI module consists of two directional operations, each implemented through a distinct sub-module: a context-aware feature fusion sub-module and a sketch-guided affine transformation sub-module, as illustrated in the bottom-right of Fig.~\ref{fig:fullnetwork}. 

In the context-aware feature fusion sub-module, the sketch feature $S_{i}$ is first processed through a zero convolution layer \cite{zhang2023adding}, and then added pixel-wise to the corresponding masked-image-encoder-modulated noisy feature $\hat{N}_{i}$. The resulting feature map, denoted as $x_{i}$, captures a coarse object contour while incorporating visual cues from the uncorrupted regions. This fused representation is subsequently passed through a $Conv$$\to$$GroupNorm$$\to$$Sigmoid$ module to predict a visual mask $ vm \in \mathbb{R}^{h_{j}\times w_{j}}$, where each pixel value ranges between $\left [ 0,1 \right ]$ and indicates the degree to which the subsequent sketch-conditional affine transformation should be applied. 

As visualized in Fig.~\ref{fig:exp_vssavisulize} (c) and (d), the intermediate output $x_{i}$ captures the fused spatial structure, while $vm$ emphasizes both the contours of uncorrupted object regions and the structural details in the partial sketch. The visual mask serves as a spatial aware mechanism that conditions the following transformation step, enabling accurate and context-aware integration of sketch features. In doing so, it establishes a coherent foundation for sketch-guided feature modulation. 

For an input batch $x_{i}\in \mathbb{R}^{b\times c\times h\times w}$, the sketch-conditional affine transformation is applied in a channel-wise manner after Group Normalization (GN), as follows:
\begin{equation}
\hat{x}_{i}^{bchw}=vm_{(h_{j},w_{j})} (\gamma_{c}(S_{i} ) \tilde{x}_{i}^{bchw}+  \beta_{c}(S_{i})), \label{eq:vssa}
\end{equation}
where $\tilde{x}_{i}^{bchw}$ is the GN-normalized features, and $\gamma_{c}(S_{i} )$ and $\beta_{c}(S_{i} )$ are two affine parameters learned from the guided sketches $S_{i}$ through the zero convolution layer. This parameterization helps reduce noise interference during the early training stages. The intermediate outputs $vm \cdot \gamma_{c}(S_{i}) $, $vm \cdot \beta_{c}(S_{i})$, and the final sketch-guided feature $\hat{SN}_{i}$ are visualized in Fig.~\ref{fig:exp_vssavisulize} (e), (f), and (g), respectively.

This sketch-conditional affine transformation modulates the influence of the partial sketch using two learnable affine parameters, while preserving sketch-guided consistency with uncorrupted object cues by additional conditioning on the visual mask predicted by the preceding fusion module. The resulting sketch-guided noisy features $\hat{SN}_{i}$ are then reintegrated into the frozen, text-embedded Stable Diffusion model, as shown in the bottom-right of Fig.~\ref{fig:fullnetwork}. This integration enables the model to inpaint partially corrupted objects with fine-grained structural guidance from the sketch, while maintaining high-level semantic coherence through the pretrained text-conditioned generative prior.

\subsection{Dataset Preparation for Model Training}\label{subsec34}

While existing datasets mainly focus on cases of complete object occlusion, they often neglect scenarios involving partial object degradation. In contrast, our method specifically addresses the inpainting of partially corrupted objects, where the occlusion mask covers a substantial portion of the object's semantic region and potentially includes parts of the background. This setting poses a non-trivial challenge, as common random masking strategies~\cite{yu2019free, liu2018image} are inadequate for such structured occlusions. These strategies typically either mask irrelevant background regions or obscure only small, semantic parts of the object, making it difficult to regulate the occlusion ratio between object and background areas. Consequently, they fail to provide reliable support for partially object-level inpainting guided by text and sketches. 

\begin{figure}[h]
  \includegraphics[width=0.48\textwidth]{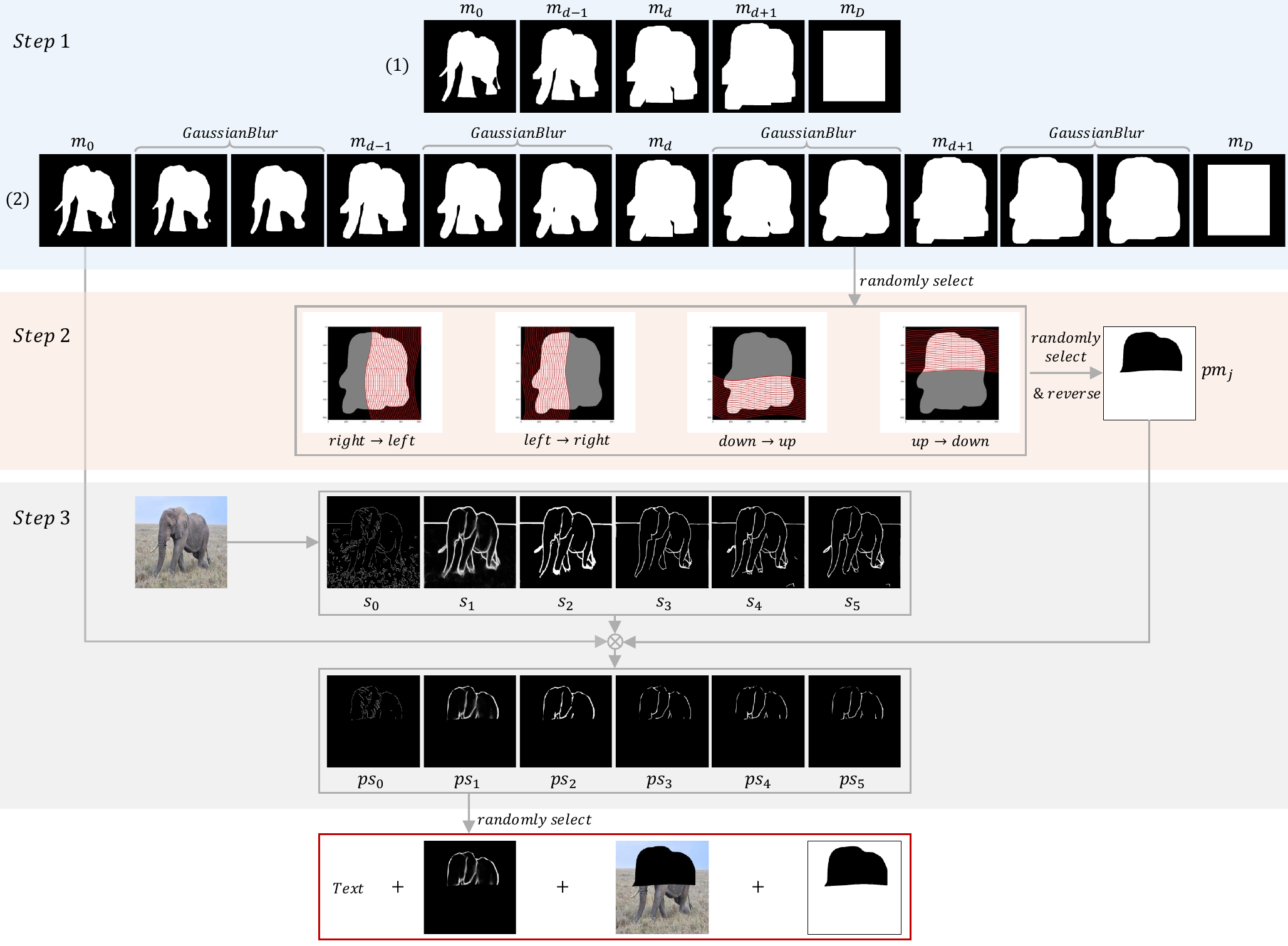}
  \setlength{\abovecaptionskip}{-1pt}
  \caption{Data preparation process for constructing partially occluded object masks and corresponding partial sketches in three steps: Step 1 (Mask Generation), where Step 1 (1) performs mask dilation to expand the object boundaries and include non-object background, and Step 1 (2) smooths the edges between adjacent dilated masks; Step 2 (Partial Masking); and Step 3 (Partial Sketch Generation). The outputs from these steps are used to construct 4-tuple data samples for each image, where the accompanying text is directly taken from the original dataset annotations.}
  \label{fig:datageneration}
\end{figure}

To overcome this limitation, we propose a tailored data preparation procedure for partially occluded object scenarios. Each generated sample comprises a partial mask, a partial sketch, a partially occluded image, and a corresponding textual description. As shown in Fig.~\ref{fig:datageneration}, we illustrate the data preparation process using a representative object image. The procedure consists of three steps: mask generation, partial masking, and partial sketch generation, which are described in detail below.

\textbf{\textit{Step 1: Mask Generation.}}
In the first step, we generate enlarged instance masks that can extend into the background regions. This is achieved using a mask dilation indicator $d\sim \left [ 0,D \right ]$, which controls the degree of dilation applied to the original instance mask $m_{0}$ provided by the annotations. The dilation process is defined as: 
\begin{equation}  
m_{d} = Dilation(m_{0}, k_{d}), \label{eq:dilation_m}
\end{equation}
where $k_{d}$ denotes the dilation kernel size. When $d=0$, the mask remains unchanged as $m_{0}$. As $d$ increases, the mask $m_{d}$ gradually expands outward. When $d=D$, the mask $m_{d}$ approximates the bounding box of the instance object, thereby losing specific shape information.

To mitigate the loss of sharp boundaries in $m_{0}$ caused by excessive dilation (as shown in Fig.~\ref{fig:datageneration} Step 1 (1)), we apply Gaussian blur between adjacent dilation masks $m_{d}$ and $m_{d+1}$ ($0\le d, d+1\le D$). This smoothing process generates masks with varying levels of spatial precision and is controlled by a mask blur indicator $s\sim \left [ 0,S \right ]$, defined as:
\begin{equation}  
m_{s} = GaussianBlur(m_{d}, m_{d+1}, k_{s}), \label{eq:blur_m}
\end{equation}
where $k_{s}$ is the Gaussian kernel size. When $s=0$, the resulting mask $m_{s}$ is identical to $m_{d}$. As $s$ increases, the mask becomes progressively smoother until $s=S$, where $m_{s}$ is equal to $m_{d+1}$. Examples of masks generated with varying $d$ and $s$ values are shown in Fig.~\ref{fig:datageneration}, Step 1 (2). One mask is randomly selected from this set for the next step.

\textbf{\textit{Step 2: Partial Masking.}}
In the second step, a B\'ezier curve is randomly generated and used to scan the selected mask from Step 1 in one of four directions: right to left, left to right, down to up, or up to down. The scanning process continues until the accumulated coverage reaches a predefined threshold (between $50\% \sim 60\%$ of the original mask area, based on our experimental settings). This operation yields four partially occluded object masks corresponding to the four scan directions. One of these masks is then randomly selected and reversed to produce the final partial mask $pm_{j}$ ($j\in \mathbb{Z}, 0\le j\le3$), as shown in Fig.~\ref{fig:datageneration}, Step 2.

\textbf{\textit{Step 3: Partial Sketch Generation.}}
In the final step, six types of sketches are generated from the clean RGB image to simulate diverse freehand drawing styles. As illustrated in Fig.~\ref{fig:datageneration}, Step 3, the sketches are created as follows: 
\begin{itemize}[label={}, leftmargin=1em]
    \item Sketch \( s_0 \): Extracted using the Canny edge detection algorithm  \cite{canny1986computational}.
    \item Sketch \( s_1 \): Generated using PidiNet \cite{su2021pixel}.
    \item Sketch \( s_2 \): Produced by applying hard-threshold filtering to \( s_1 \).
    \item Simplified Sketches \( s_3, s_4, s_5 \): Created using the rough sketch simplification (RSS) algorithm \cite{mo2021general}, applied to \( s_1 \) with three different initialization strokes.
\end{itemize}

The partial sketch $ps_{i}$ is calculated using the formula $(1-pm_{j})\otimes m_{0}\otimes s_{i}$, where $i\in \mathbb{Z}$ and $0\le i\le5$, and $m_{0}$ represents the original instance mask. This operation extracts the partial sketch corresponding to the object within the partial mask $pm_{j}$. The goal of this step is to enrich the training set with a variety of sketch types, thereby improving the model's generalization to user-drawn sketches during inference. Finally, one partial sketch is randomly selected and incorporated into the 4-tuple data sample.

Using the above process with annotations from CUB \cite{wah2011caltech} and MSCOCO \cite{lin2014microsoft}, we construct two customized datasets: CUB-sketch and MSCOCO-sketch, which serve as benchmarks for sketch-guided partial object inpainting.  

\section{Experiments}\label{sec:experiments}
We evaluate the effectiveness of the proposed method against state-of-the-art diffusion-based approaches. Extensive experimental results demonstrate the superiority of our pipeline in achieving fine-grained posture guidance using partial sketches for inpainting partially corrupted objects, while preserving sketch-guided consistency with uncorrupted object cues. Finally, we perform ablation studies to assess the contribution of each component within the proposed framework to the overall performance.
\begin{table*}[!htbp]
\centering
\setlength{\abovecaptionskip}{1pt}
\caption{Quantitative comparisons with state-of-the-art diffusion-based methods for \textbf{partially corrupted object inpainting} on the test sets of CUB-sketch and MSCOCO-sketch. $\uparrow$ indicates higher is better, and $\downarrow$ indicates lower is better. The best scores are marked in bold.}
\label{tab:table1}
\centering
\setlength{\tabcolsep}{8pt} 
\renewcommand{\arraystretch}{1.2} 
\resizebox{0.98\textwidth}{!}{%
\footnotesize
\begin{tabular}{l|cccc|cccc}
\bottomrule
\multirow{2}{*}{Method}              & \multicolumn{4}{c|}{CUB-sketch}   & \multicolumn{4}{c}{MSCOCO-sketch} \\
              & AS $\uparrow$   & CLIP Score $\uparrow$ & FID $\downarrow$ & LPIPS ({\scriptsize$\times 10^{2}$}) $\downarrow$ & AS $\uparrow$   & CLIP Score $\uparrow$  & FID $\downarrow$ & LPIPS ({\scriptsize$\times 10^{2}$}) $\downarrow$ \\ \hline
SD-Inpainting \cite{rombach2022high} & 5.77 & 29.01      & 8.21 & 10.01 & 5.64  & 25.29       & 4.87 & 12.06\\ \hline
BrushNet \cite{ju2024brushnet}      & 5.76 & 29.02      & 9.79 & 11.22 & 5.68  & 25.35       & 5.56 & 13.99\\ \hline
PowerPaint \cite{zhuang2025task}   & 5.71 & 28.52      & 10.05 & 12.09 & 5.65  & 25.09       & 5.06 & 13.92\\ \hline
MaGIC \cite{wang2024magic}        & 5.79 & 28.67      & 8.83 & 8.72 & 5.59  & 25.78       & 4.90  & 11.52\\ \hline
ControlNet \cite{zhang2023adding}        & 5.73 & 28.53      & 10.77 & 12.01 & 5.61  & 25.03       & 5.22 & 13.84 \\ \hline
PowerPaint-ControlNet \cite{zhuang2025task}        & 5.73 & 29.03      & 8.78 & 12.79 & 5.63  & 25.46       & 4.89 & 15.72 \\ \hline
Ours          & \textbf{5.81} & \textbf{29.08}      & \textbf{8.17}  & \textbf{8.48} & \textbf{5.71}  & \textbf{25.90}       & \textbf{4.83} & \textbf{10.95} \\ \toprule
\end{tabular}}
\vspace{-2pt}
\end{table*}
\subsection{Datasets}\label{subsec41}
We train the proposed pipeline using four-tuple data consisting of partially masked images, partial masks, partial sketches, and corresponding text descriptions. Existing diffusion-based inpainting methods guided by text or sketches do not specifically target the restoration of partially occluded objects. Moreover, the datasets they commonly use, such as CUB \cite{wah2011caltech} and MSCOCO \cite{lin2014microsoft}, lack the necessary partial masks and partial sketches required for this task. To generate the four-tuple data, we leverage the instance masks and text captions from the original datasets to construct two new datasets: CUB-sketch and MSCOCO-sketch, as described in Sec.~\ref{subsec34}. 

The resulting CUB-sketch and MSCOCO-sketch datasets contain 11,788 and 30,845 four-tuple samples, respectively. These datasets, along with the associated code, are available at \url{https://github.com/yonglezhang95/PartiallyObjectInpainting}. Specifically, CUB-sketch includes 8,855 samples for training and 2,933 for testing, while MSCOCO-sketch comprises 20,526 training samples and 10,283 testing samples. The MSCOCO-sketch training set is derived from the MSCOCO training set, and its testing set is sourced from the MSCOCO validation set.

\subsection{Experimental Setup}\label{subsec42}
\subsubsection{\textbf{\textit{Compared Methods}}}\label{subsub421}We compare our pipeline with six state-of-the-art diffusion-based methods: SD-Inpainting \cite{rombach2022high}, BrushNet \cite{ju2024brushnet}, PowerPaint \cite{zhuang2025task}, MaGIC \cite{wang2024magic}, ControlNet \cite{zhang2023adding}, and PowerPaint-ControlNet \cite{zhuang2025task}, the latter of which integrates the ControlNet adapter. The first three methods are text-guided inpainting approaches, while the latter three are sketch-guided inpainting methods based on the pretrained text-to-image Stable Diffusion model.

To ensure fair comparison, we adopt the recommended hyperparameters provided in each method's official implementation and apply them consistently to our training and testing datasets. For SD-Inpainting, BrushNet, and PowerPaint, we use 3-tuple data samples comprising partial masks, partially masked images, and text prompts. For MaGIC, ControlNet, PowerPaint-ControlNet, and our proposed pipeline, we use 4-tuple data samples consisting of partial masks, partially masked images, text prompts, and partial sketches. All methods perform inference using 50 diffusion steps and a Classifier-Free Guidance (CFG) scale of 7.5, which is a standard setting in prior works~\cite{ho2021classifier, rombach2022high, zhuang2025task}. 

\subsubsection{\textbf{\textit{Evaluation Metrics}}}\label{subsub422}We evaluate all methods using four widely adopted metrics: Aesthetic Score (AS)~\cite{kao2015visual, schuhmann2022laion}, Fr\'echet Inception Distance (FID)~\cite{heusel2017gans}, CLIP Score~\cite{hessel2021clipscore}, and Learned Perceptual Image Patch Similarity (LPIPS)~\cite{zhang2018unreasonable}. AS quantifies image quality based on human perception using a linear regression model trained on real image rating pairs. FID assesses the realism and visual consistency of the recovered object within the overall image distribution. CLIP Score evaluates the semantic alignment between the generated content and the input text prompt. LPIPS measures perceptual similarity by comparing deep features extracted from pretrained neural networks, providing a reliable estimate of how close the restored image is to the ground truth. 

In addition to these quantitative metrics, we conduct a human study to assess the alignment between the guided sketch and the restored object within the damaged region. This evaluation reflects how well the model respects the sketch guidance and maintains visual consistency with the uncorrupted object regions during the inpainting process, which is difficult to capture through automated metrics alone.

\subsubsection{\textbf{\textit{Parameter Setting}}}\label{subsub423}We use the Adam optimizer with a learning rate of 0.00001 and a batch size of 16 to train our pipeline, based on the pre-trained Stable Diffusion v1.5. During training, to avoid overfitting, we sample partial masks, segmentation masks, and bounding box masks with a probability ratio of 6:3:1.
\begin{figure*}[]
  \centering
  \setlength{\abovecaptionskip}{-1pt}
  \includegraphics[width=0.98\textwidth]{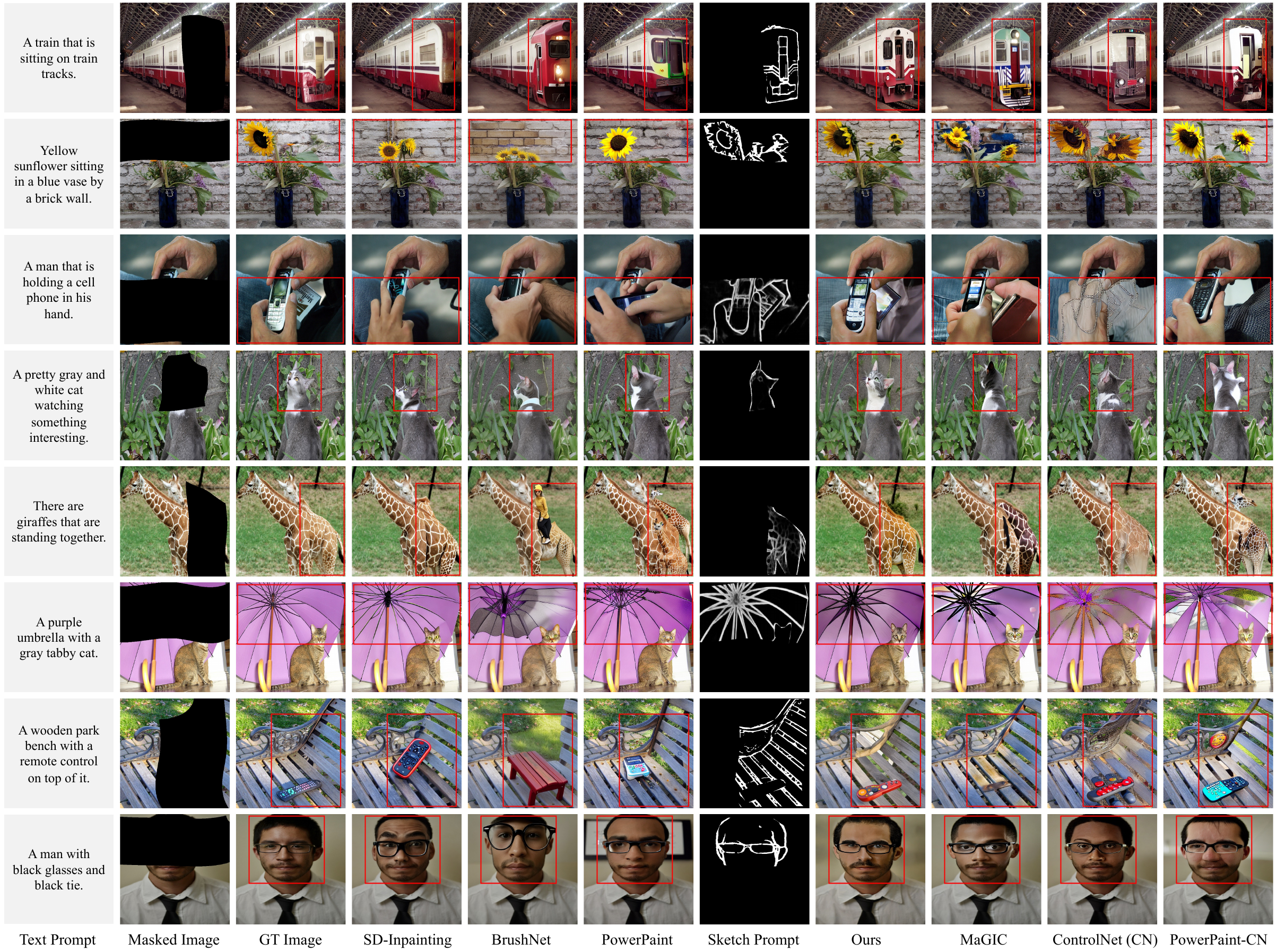}
  \caption{Qualitative comparison of our method with SD-Inpainting \cite{rombach2022high}, BrushNet \cite{ju2024brushnet}, PowerPaint \cite{zhuang2025task}, MaGIC \cite{wang2024magic}, ControlNet \cite{zhang2023adding}, and PowerPaint-ControlNet \cite{zhuang2025task} for \textbf{partially corrupted object inpainting} on the CUB-sketch and MSCOCO-sketch test images. Among these methods, MaGIC, ControlNet, and PowerPaint-ControlNet utilize both text and sketch prompts, while the other three rely solely on the text prompt.}
  \label{fig:exp_fig3}
\end{figure*}
\subsection{Experimental Comparisons}\label{sec43}
\subsubsection{\textbf{\textit{Quantitative Results}}}\label{sec431} Table~\ref{tab:table1} presents the numerical results of all methods evaluated on the CUB-sketch and MSCOCO-sketch test sets. The results across four metrics—AS, CLIP Score, FID, and LPIPS—consistently demonstrate the effectiveness of our pipeline in restoring partially corrupted objects.

On the CUB-sketch test set, PowerPaint and ControlNet exhibit inferior performance across all four metrics. This can be attributed to their limited ability to model contextual relationships between uncorrupted objects and corrupted regions. Specifically, ControlNet's inherent controllability, derived from Text-to-Image (T2I) tasks, makes it less effective for object restoration under partial occlusions, as it injects sketches independently, disregarding the ongoing inpainting process. Similarly, MaGIC's performance is surpassed by our method, which leverages multi-scale bidirectional feature interaction mechanisms. These mechanisms dynamically weight sketch guidance while adapting to uncorrupted object regions, mitigating the instability of MaGIC's indirect, gradient-based guidance. As a result, our approach achieves superior restoration outcomes. 

On the MSCOCO-sketch test set, methods such as SD-Inpainting and BrushNet, which rely solely on text prompts without spatial sketch control, underperform compared to our pipeline. Our approach utilizes four-tuple inputs (corrupted image, text, partial sketch, and mask) integrated with a frozen text-embedded Stable Diffusion model, enabling finer object pose details through sketch guidance and pretrained textual semantic priors. In contrast, MaGIC, ControlNet, and PowerPaint-ControlNet produce suboptimal results, likely due to their lack of specialized modules for perceiving visible object contexts and facilitating tailored restoration of partially corrupted scenes.

Additionally, Table~\ref{tab:table3} compares the performance of our method with existing diffusion-based approaches for fully corrupted object restoration on the CUB-sketch and MSCOCO-sketch test sets. In these cases, objects are entirely masked using either instance segmentation masks or bounding box masks, with a ratio of 8:2 in the test set. We introduce a Local-LPIPS (L-LPIPS) metric to measure perceptual similarity between the fully inpainted object and the ground truth instance. Among the compared methods, our approach achieves competitive qualitative metrics, further validating that bidirectional interaction between guidance information and inpainting features during the denoising process enhances restoration performance compared to independent integration of guidance information.

\begin{table*}[!htbp]
\centering
\setlength{\abovecaptionskip}{1pt}
\caption{Quantitative comparisons with state-of-the-art diffusion-based methods for \textbf{fully corrupted object inpainting} on the test sets of CUB-sketch and MSCOCO-sketch. $\uparrow$ indicates higher is better, and $\downarrow$ indicates lower is better. The best scores are marked in bold.}
\label{tab:table3}
\centering
\setlength{\tabcolsep}{2pt} 
\renewcommand{\arraystretch}{1.2} 
\resizebox{0.98\textwidth}{!}{%
\footnotesize
\begin{tabular}{l|ccccc|ccccc}
\bottomrule
\multirow{2}{*}{Method}              & \multicolumn{5}{c|}{CUB-sketch}   & \multicolumn{5}{c}{MSCOCO-sketch} \\
              & AS $\uparrow$   & CLIP Score $\uparrow$ & FID $\downarrow$ & L-LPIPS ({\scriptsize$\times 10^{2}$}) $\downarrow$ & LPIPS ({\scriptsize$\times 10^{2}$}) $\downarrow$ & AS $\uparrow$   & CLIP Score $\uparrow$  & FID $\downarrow$ & L-LPIPS ({\scriptsize$\times 10^{2}$}) $\downarrow$ & LPIPS ({\scriptsize$\times 10^{2}$}) $\downarrow$ \\ \hline
SD-Inpainting \cite{rombach2022high} & 5.65 & 28.13      & 17.50 & 7.68 & 13.50  & 5.70       & 25.81 & 7.53 & 10.66 & 15.51 \\ \hline
BrushNet \cite{ju2024brushnet}      & 5.90 & 28.64      & 19.21 & 7.90 & 14.18  & 5.68       & 25.85 & 9.24 & 11.54 & 17.48 \\ \hline
PowerPaint \cite{zhuang2025task}   & 5.82 & 28.24      & 18.38 & 7.44 & 12.85  & 5.66       & 25.72 & 8.38 & 11.82 & 17.24 \\ \hline
MaGIC \cite{wang2024magic}        & \textbf{5.94} & 28.88      & 17.96 & 6.54 & \textbf{10.85}  & 5.78       & 25.83  & \textbf{7.21} & 10.19 & \textbf{14.93}\\ \hline
ControlNet \cite{zhang2023adding}        & 5.78 & 28.51      & 16.95 & 6.36 & 13.32  & 5.71       & 25.78 & 7.46 & 11.09 & 16.72  \\ \hline
PowerPaint-ControlNet \cite{zhuang2025task}        & 5.81 & 28.27      & 18.02 & 7.23 & 14.06  & 5.65       & 25.73 & 8.01 & 11.36 & 18.59 \\ \hline
Ours          & \textbf{5.94} & \textbf{29.01}      & \textbf{16.74}  & \textbf{6.33} & 10.89  & \textbf{5.80}       & \textbf{25.99} & 7.48 & \textbf{9.97} & 15.35 \\ \toprule
\end{tabular}}
\vspace{-2pt}
\end{table*}
\begin{figure*}[htbp]
  \centering
  \setlength{\abovecaptionskip}{-1pt}
  \includegraphics[width=0.98\textwidth]{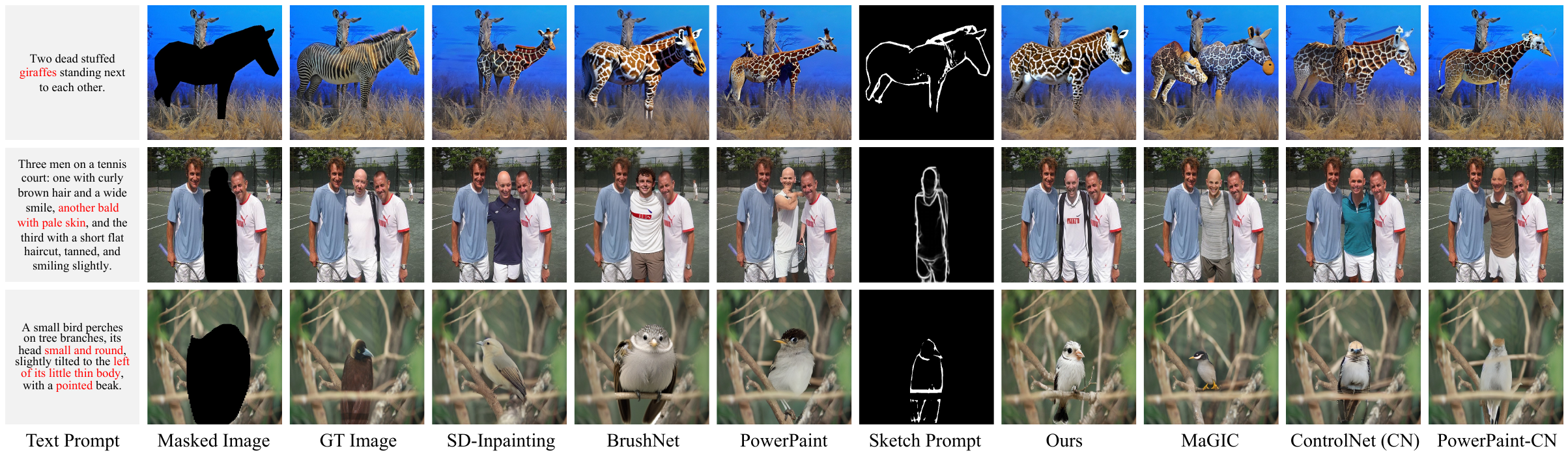}
  \caption{Qualitative comparison of our method with SD-Inpainting \cite{rombach2022high}, BrushNet \cite{ju2024brushnet}, PowerPaint \cite{zhuang2025task}, MaGIC \cite{wang2024magic}, ControlNet \cite{zhang2023adding}, and PowerPaint-ControlNet \cite{zhuang2025task} for \textbf{fully corrupted object inpainting} on the CUB-sketch and MSCOCO-sketch test images. Among these methods, MaGIC, ControlNet, and PowerPaint-ControlNet utilize both text and sketch prompts, while the other three rely solely on the text prompt.}
  \label{fig:mainexp_segmask}
\end{figure*}
\subsubsection{\textbf{\textit{Qualitative Results}}}\label{sec432} Fig.~\ref{fig:exp_fig3} presents qualitative results for six comparison methods. Our pipeline generates high-fidelity visual details in restored objects, achieving strong alignment with both the sketch prompt and textual semantics.

Methods relying solely on text prompts, such as SD-Inpainting, BrushNet, and PowerPaint, produce semantically meaningful object restorations but struggle with visual consistency and exhibit arbitrary spatial postures. For instance, in Fig.~\ref{fig:exp_fig3}, the restored train heads (first row), hands (third row), and umbrellas (sixth row) display inconsistent spatial structures relative to uncorrupted object regions. These methods often fail to control object poses in inpainted regions, highlighting the limitations of text-based guidance in achieving fine-grained posture accuracy.

In contrast, MaGIC, ControlNet, and PowerPaint-CN incorporate the same sketch prompt as our method but produce inpainted results with inconsistent object postures relative to the sketch. For example, the sunflower shape (second row) and the man's hand (third row) in Fig.~\ref{fig:exp_fig3} reveal difficulties in controlling the completion of partially corrupted objects using sketch prompts. Our method, however, integrates a masked image encoder to identify corrupted regions and uncorrupted object contexts, a sketch-conditional encoder with multi-scale bidirectional feature interaction modules to ensure precise sketch integration and consistency with uncorrupted regions, and a pretrained Stable Diffusion model for robust text-guided priors. This results in high-fidelity object completion with both visual and semantic coherence.

Fig.~\ref{fig:mainexp_segmask} provides additional qualitative comparisons for fully corrupted object inpainting. Our method consistently outperforms existing approaches, demonstrating superior visual alignment with the guiding sketch prompt (e.g., the restored bird in the third row) and improved consistency with undamaged contexts. Text-guided methods, such as SD-Inpainting, BrushNet, and PowerPaint, fail to achieve pixel-level accuracy in object postures, even when provided with detailed text prompts, such as ``its head slightly tilted to the left of its little thin body'' for the bird inpainting example (third row). For instance, the inpainted bird's head in SD-Inpainting and PowerPaint tilts to the right, contrary to the prompt's description.

\subsection{Comparison Between Text-Only and Text+Sketch Guidance}\label{sec44}
In Sec.~\ref{sec43}, we demonstrated that text prompts alone are insufficient for guiding diffusion-based models to achieve pixel-level accuracy in object pose inpainting. Here, we further compare the visual outcomes of inpainting guided solely by text prompts with those guided by both text and sketch prompts, focusing on our method, MaGIC, and ControlNet.

\begin{figure}[t]
  \centering
  \setlength{\abovecaptionskip}{-1pt}
  \includegraphics[width=0.48\textwidth]{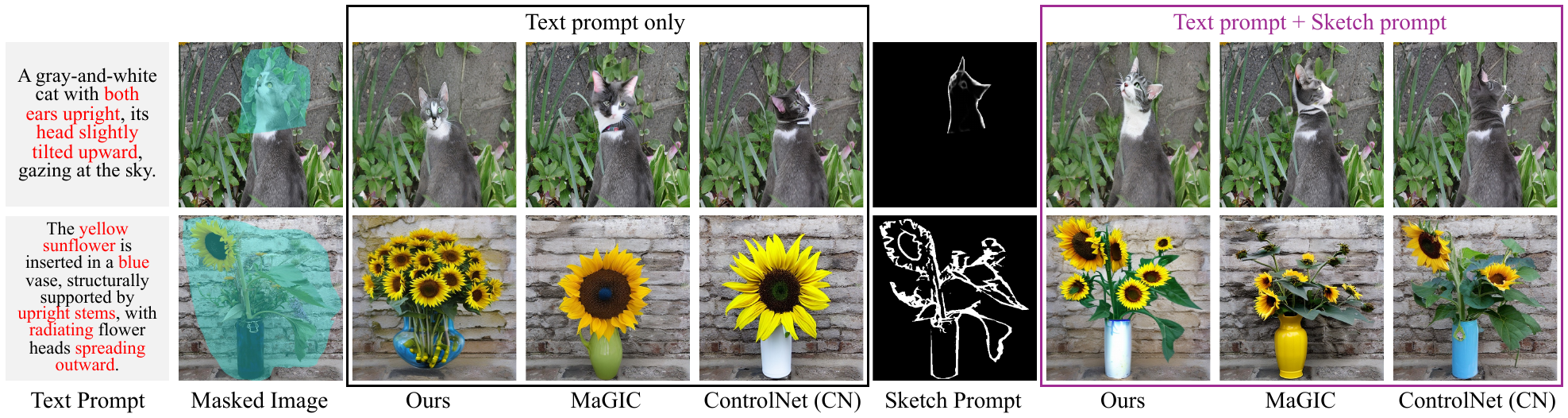}
  \caption{Qualitative comparison of our method with MaGIC \cite{wang2024magic} and ControlNet \cite{zhang2023adding} for object inpainting under text-only guidance and combined text and sketch prompts.}
  \label{fig:exp_textorsketch}
\end{figure}

In the text-only setting, the sketch prompt is replaced with a completely black image. The left side of Fig.~\ref{fig:exp_textorsketch} displays the inpainting results for each method using only text prompts. For example, in the first row, ControlNet fails to capture the structural description ``both ears upright'' from the text prompt. Similarly, in the second row, both MaGIC and ControlNet fail to depict the ``upright stems'' specified in the text. Although our method also struggles to fully reconstruct the spatial structure implied by the text, these results highlight the inherent limitations of text prompts, which provide high-level semantic cues but lack pixel-level spatial specificity.

When sketch prompts are incorporated, as shown on the right side of Fig.~\ref{fig:exp_textorsketch}, our method achieves significantly more accurate pixel-level pose reconstructions, closely aligning with the guided sketch structure compared to MaGIC and ControlNet. This improvement arises from our sketch-guided bidirectional feature interaction mechanism, which ensures consistency between the sketch guidance and the inpainting process by adapting to the surrounding object context—a capability absent in MaGIC and ControlNet.

\subsection{Subjective Assessment via User Study}\label{sec45}
We conducted two user studies to evaluate user preference and sketch alignment score metrics. For the user preference assessment, we randomly selected 35 partially corrupted object images from the MSCOCO-sketch test set and inpainted them using six competing methods and our proposed pipeline. Twenty participants were recruited, half of whom were image processing majors, while the other half were PhD students from other disciplines. Each participant was presented with the corrupted images, their corresponding inpainted results, and text prompts, and asked to select one or more inpainted images they perceived as natural with high semantic fidelity. As shown in the User Preference column of Table~\ref{tab:table4}, our method's inpainted results were preferred in $59.33\%$ of the selections. 

For the sketch alignment score assessment, we randomly selected 15 partially corrupted object images from the MSCOCO-sketch test set, each accompanied by four partial sketches at different scales as guides. Twenty participants were presented with inpainting results from competing sketch-guided methods and our pipeline, along with the guiding sketches, and asked to rate how well the inpainted object regions aligned with the sketch and remained consistent with the uncorrupted regions, using a score from 1 (poor alignment) to 5 (best alignment). The Sketch Alignment Score column in Table~\ref{tab:table4} indicates that participants favored the inpainting results of our method.
\begin{table}[h]
\centering
\footnotesize
\setlength{\abovecaptionskip}{1pt}
\caption{User Study Results for User Preference and Sketch Alignment Score. The study involved 20 participants. User Preference assesses the naturalness and textual semantic fidelity of inpainted object images, while Sketch Alignment Score evaluates the alignment of inpainted object regions with the guiding sketch and their consistency with uncorrupted regions. $\uparrow$ indicates higher is better.}
\label{tab:table4}
\renewcommand{\arraystretch}{1.3}
\resizebox{1.0\linewidth}{!}{
\begin{tabular}{l|c|c}
\toprule
Method & User Preference ({\scriptsize $\%$}) $\uparrow$ & \begin{tabular}[c]{@{}c@{}}Sketch Alignment Score $\uparrow$\\  (1 = poor, 5 = best)\end{tabular} \\ \hline
SD-Inpainting \cite{rombach2022high}     & 9.83                  & \text{--}                                                                                               \\ \hline
BrushNet \cite{ju2024brushnet}        & 23.33                  & \text{--}                                                                                               \\ \hline
PowerPaint \cite{zhuang2025task}       & 10.33                  & \text{--}                                                                                               \\ \hline
MaGIC \cite{wang2024magic}       & 14.66                  & 3.7366                                                                                        \\ \hline
ControlNet \cite{zhang2023adding}       & 9.83                  & 3.7033                                                                                               \\ \hline
PowerPaint-ControlNet \cite{zhuang2025task}       & 10.66                  & 2.7133                                                                                               \\ \hline
Ours      & \textbf{59.33}                  & \textbf{4.0700}                                                                                               \\ \bottomrule
\end{tabular}}
\end{table}

\subsection{Model Flexibility with Diverse Text Prompts and Various Sketches}\label{sec46}
\begin{figure}[h]
  \centering
  \setlength{\abovecaptionskip}{-1pt}
  \includegraphics[width=0.48\textwidth]{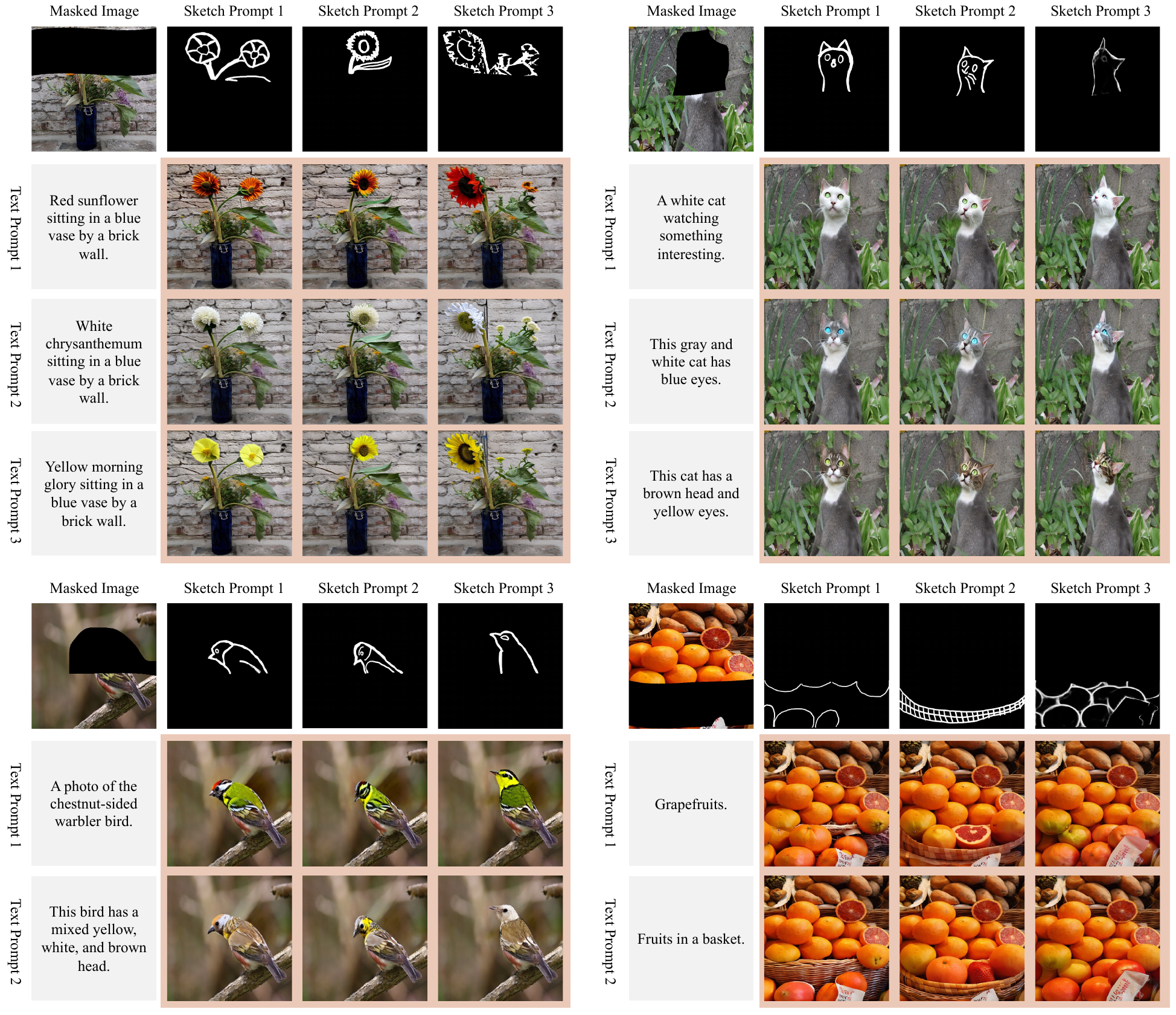}
  \caption{Controllable object inpainting results generated by our pipeline, conditioned on different combinations of sketch and text prompts, demonstrating high-fidelity outcomes.}
  \label{fig:exp_fig4}
\end{figure}

Fig.~\ref{Figure:threetypesofinpainting} (c)$\sim$(e) present the results generated by our pipeline using the same text prompt but different sketch prompts. Fig.~\ref{fig:exp_fig4} showcases more controllable object inpainting, conditioned on various combinations of sketch and text prompts. Our pipeline produces high-fidelity inpainted results that maintain visual consistency with the sketch prompt and semantic consistency with the text prompt.

\begin{figure*}[h]
  \centering
  \includegraphics[width=0.94\textwidth]{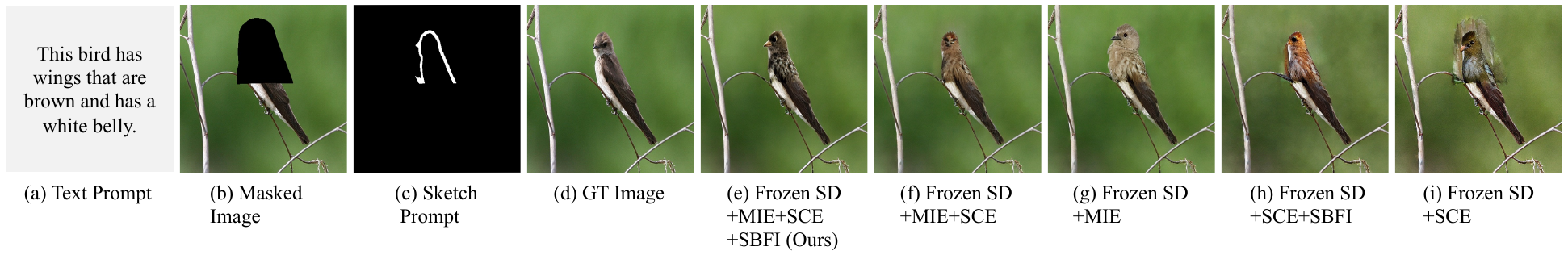}
  \setlength{\abovecaptionskip}{-1pt}
  \caption{Qualitative comparisons from ablation studies on different component configurations in our pipeline. Frozen SD denotes the pre-trained text-to-image Stable Diffusion model with parameters frozen. Each panel shows the generated image under different model variants to illustrate the effect of each component.}
  \label{fig:exp_fig2}
\end{figure*}

\section{Ablation Studies}\label{sec:ablation}
We conducted ablation studies to assess the effect of different component designs in our pipeline, which comprises three main components: the Masked Image Encoder (MIE), the Sketch-Conditional Encoder (SCE), and the Sketch-guided Bidirectional Feature Interaction (SBFI) module. We verify corresponding effects by examining combinations of these components at each stage. We evaluated their effects by testing different combinations of these components at each stage. Table~\ref{tab:table2} presents the numerical results of these models on the CUB-sketch test set for partially corrupted object inpainting.

In Table~\ref{tab:table2}, models that include the SCE but exclude the SBFI module integrate sketch information into the Frozen Stable Diffusion model via element-wise addition, replacing the SBFI module. Similarly, for models with the SCE but without the MIE, we expanded the input channel dimension of the Stable Diffusion model to incorporate the masked image and mask. All ablation models were trained under the same training configuration as our full pipeline.
\begin{table}[!htbp]
\centering
\footnotesize
\setlength{\abovecaptionskip}{1pt}
\caption{\footnotesize Quantitative comparisons from ablation studies on different component designs in our pipeline. $\uparrow$ indicates higher is better, and $\downarrow$ indicates lower is better.}
\label{tab:table2}
\renewcommand{\arraystretch}{1.4} 
\resizebox{1.0\linewidth}{!}{%
\begin{tabular}{l|ccc}
\toprule
        \multirow{2}{*}{Model}                     & \multicolumn{3}{c}{CUB-sketch}   \\
                             & AS $\uparrow$   & CLIP Score $\uparrow$ & FID $\downarrow$   \\ \hline
Frozen SD+SCE                & 5.46 & 27.65      & 13.94 \\ \hline
Frozen SD+SCE+SBFI           & 5.54 & 27.85      & 10.88 \\ \hline
Frozen SD+MIE            & 5.76 & 28.92      & 9.91  \\ \hline
Frozen SD+MIE+SCE            & 5.78 & 29.03      & 8.48  \\ \hline
Frozen SD+MIE+SCE+SBFI(Ours) & \textbf{5.81} & \textbf{29.08}      & \textbf{8.17}  \\ \bottomrule
\end{tabular}}
\end{table}

\subsection{Effect of Sketch-guided Bidirectional Feature Interaction (SBFI).}\label{subsubsec5.1} 
The quantitative results in Table~\ref{tab:table2} demonstrate that incorporating the SBFI module significantly improves model performance. Specifically, the model ``Frozen SD + SCE'' underperforms compared to ``Frozen SD + SCE + SBFI'' in terms of AS, CLIP Score, and FID. Similarly, the model ``Frozen SD + MIE + SCE'' shows inferior performance than ``Frozen SD + MIE + SCE + SBFI (Ours).'' The qualitative comparisons in Fig.~\ref{fig:exp_fig2} further support these findings: the model ``Frozen SD + MIE + SCE + SBFI (Ours)'' generates a sharper and more accurate outline of the bird's head than ``Frozen SD + MIE + SCE,'' and more visual artifacts appear in the corrupted object region produced by the model ``Frozen SD + SCE" compared to ``Frozen SD + SCE + SBFI.'' 

These quantitative and qualitative results indicate that the SBFI module effectively incorporates sketch prompt information into the object restoration process while preserving visual consistency with the uncorrupted parts of the object. The performance gain is attributed to the module's ability to fuse multi-scale sketch-derived and corrupted-object-modulated features, ensuring spatial alignment for subsequent sketch integration based on affine transformation.

\subsection{Effect of Masked Image Encoder (MIE)}\label{subsubsec5.2} 
As shown in Table~\ref{tab:table2}, the model ``Frozen SD + MIE + SCE'' achieves better results than ``Frozen SD + SCE'' across all three evaluation metrics. Likewise, ``Frozen SD + MIE + SCE + SBFI'' outperforms ``Frozen SD + SCE + SBFI.'' These findings suggest that the MIE plays a crucial role in generating an initial visual representation that clearly distinguishes between corrupted and uncorrupted regions. This region-aware information provides multi-scale spatial object context and mask localization, enabling the sketch prompt to be better adapted to the uncorrupted object content. The differences in visual quality are further illustrated in Fig.~\ref{fig:exp_fig2}, panels (f) and (i), and (e) and (h), respectively.

\subsection{Effect of Sketch-Conditional Encoder (SCE)}\label{subsubsec5.3} 
The fourth and fifth columns of Table~\ref{tab:table2} show that adding the SCE component to the model ``Frozen SD + MIE'' leads to performance improvements. The SCE extracts multi-scale sketch features, learning different levels of representative information from the input sketch. Furthermore, incorporating the SBFI module into ``Frozen SD + MIE + SCE'' yields even better results, as seen in the final (sixth) row of Table~\ref{tab:table2}. This underscores the importance of effectively integrating sketch information into the corrupted object regions while maintaining consistency with the uncorrupted parts—both of which are well achieved by the proposed SBFI module. 

Qualitative results in Fig.~\ref{fig:exp_fig2}, panels (f) and (g), show that without the SCE module, the model ``Frozen SD + MIE'' generates a larger bird head with an unstable posture. In contrast, the full model ``Frozen SD + MIE + SCE + SBFI'' demonstrates finer control over the object’s shape and posture than ``Frozen SD + MIE + SCE'', even when guided by a simplified hand-drawn sketch, as shown in Fig.~\ref{fig:exp_fig2}, panels (e) and (f).

\subsection{Inpainting Guided by Prompts from Abstract to Detailed}\label{subsubsec5.4} 
\begin{figure}[h]
  \centering
  \setlength{\abovecaptionskip}{-1pt}
  \includegraphics[width=0.3\textwidth]{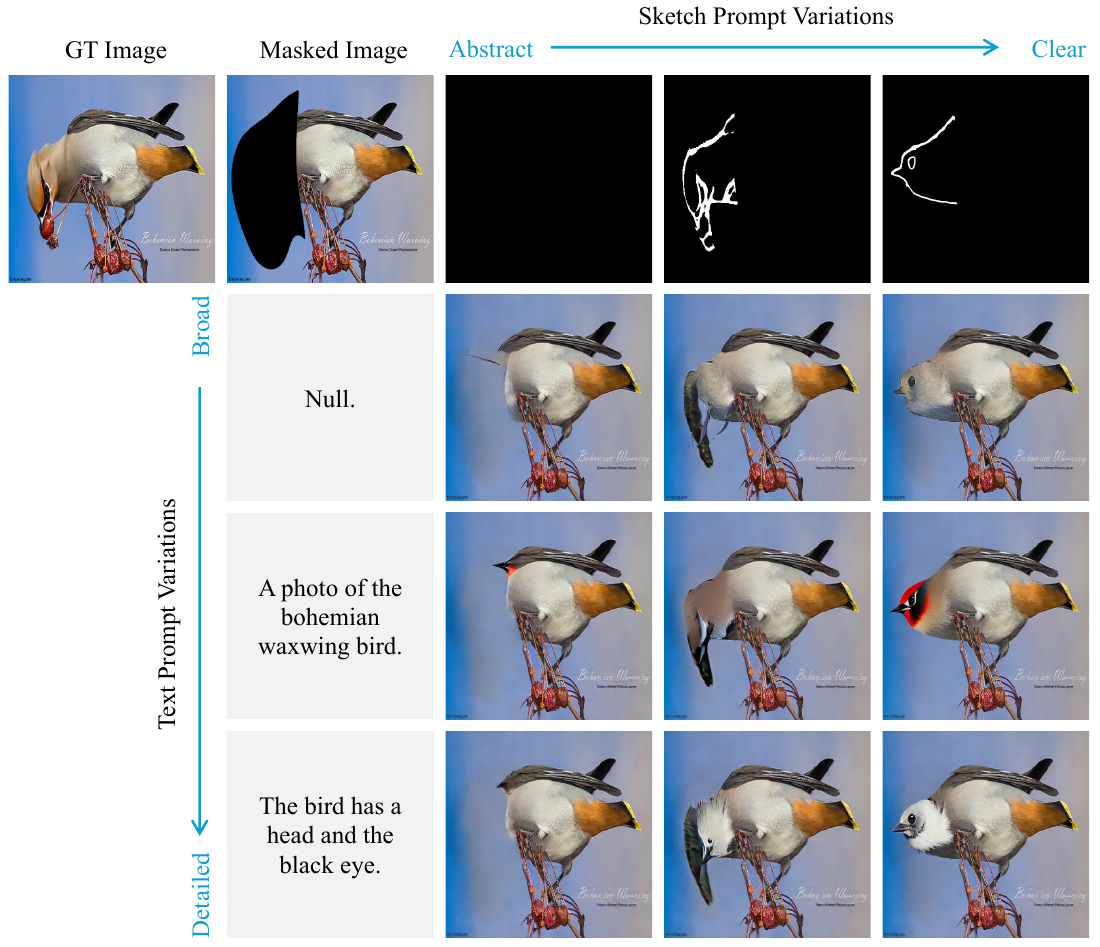}
  \caption{Object inpainting results from our pipeline under varying levels of sketch and text prompt specificity. Sketch prompts range from abstract (a completely black input with no sketch) to clear (a precise outline of a bird's head). Text prompts range from broad (an empty prompt) to detailed (a well-defined text prompt describing the bird's head).}
  \label{fig:promptsfromAbstracttoDetailed}
\end{figure}

The object inpainting results in Fig.~\ref{fig:promptsfromAbstracttoDetailed} reveal a key trend: as the sketch prompt transitions from abstract to clear and the text prompt shifts from a broad to a more detailed description of the corrupted area, the inpainted objects exhibit increased visual and semantic consistency.

\subsection{Limitations}\label{subsubsec5.5} 
\begin{figure}[h]
  \centering
  \setlength{\abovecaptionskip}{-1pt}
  \includegraphics[width=0.48\textwidth]{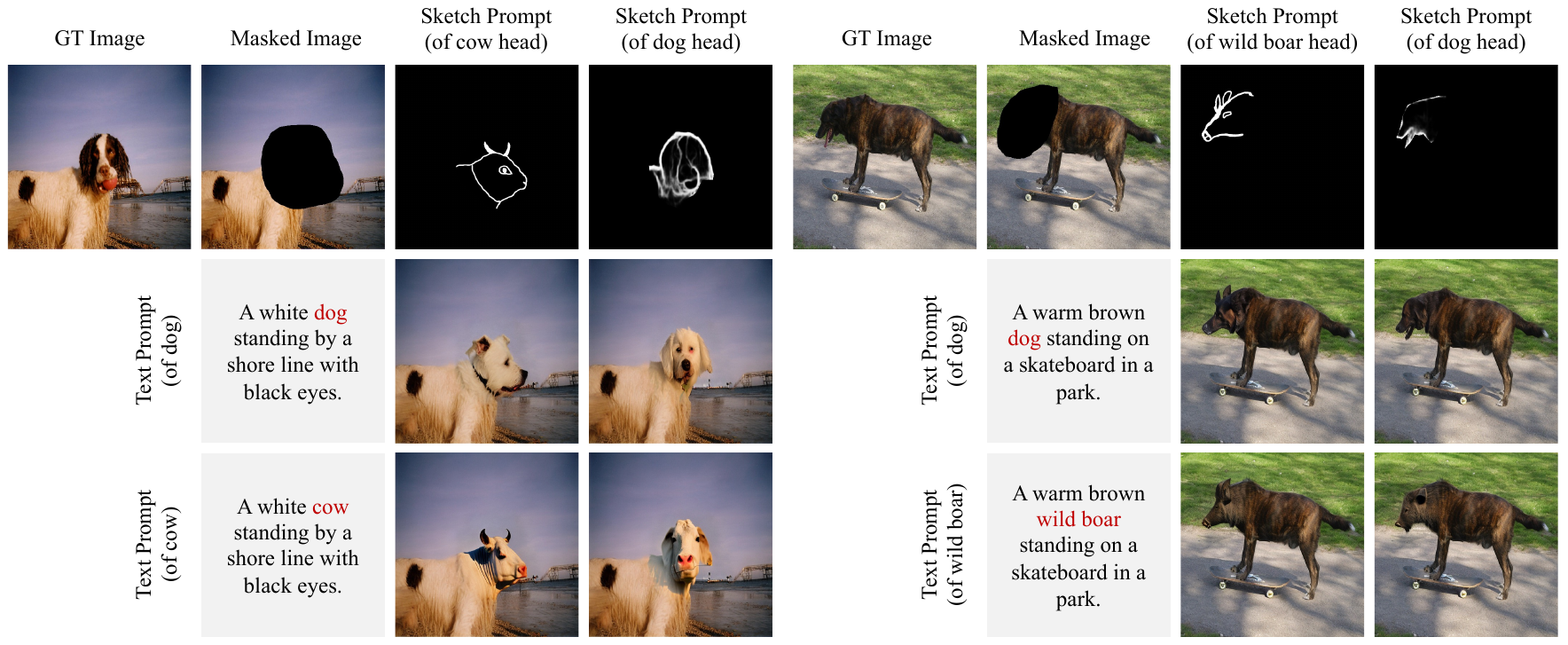}
  \caption{Object inpainting results generated by our pipeline in scenarios where the sketch prompt is inconsistent with the text prompt. For instance, the sketch may depict a cow's head, while the text prompt instead describes a different object, such as a dog, leading to incoherent completions.}
  \label{fig:conflictingSketchandText}
\end{figure}

We examine scenarios where the content of the guiding sketch is inconsistent with or contradicts the description provided by the text prompt. For example, if the sketch depicts a cow's head while the text prompt describes a different object, such as a dog, our pipeline struggles to produce a coherent completion, as illustrated in Fig.~\ref{fig:conflictingSketchandText}. Such inconsistencies lead the model to generate unrealistic and incoherent results.

\begin{figure}[h]
  \centering
  \setlength{\abovecaptionskip}{-1pt}
  \includegraphics[width=0.48\textwidth]{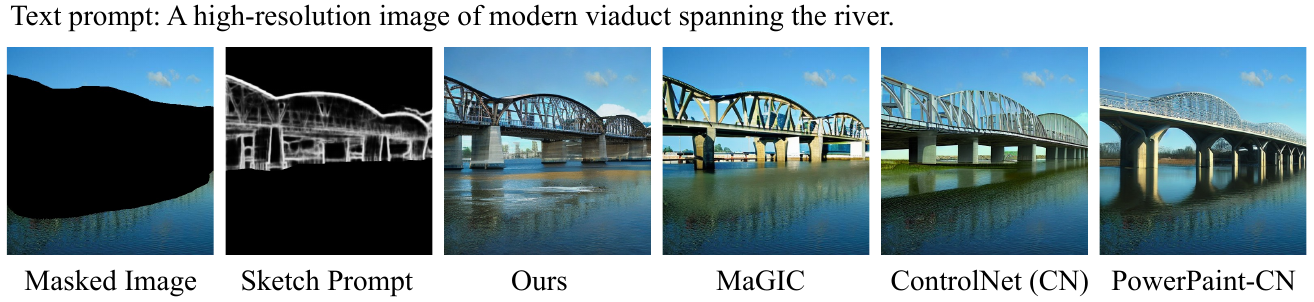}
  \caption{Inpainting result using a complex and noisy sketch prompt to guide the reconstruction of a viaduct, where these methods fail to recover a clear structure.}
  \label{fig:exp_fuzzy_sketch}
\end{figure}

In addition, when the sketch prompt contains a complex structure with noise or ambiguity to guide object inpainting, as in the sketch shown in Fig.~\ref{fig:exp_fuzzy_sketch}, our method fails to reconstruct a clear structural layout of the viaduct, even though the result still outperforms MaGIC, ControlNet, and PowerPaint-CN to some extent. This limitation arises because our method, just as with MaGIC, ControlNet, and PowerPaint-CN, lacks the ability to correct poor-quality sketch inputs.

\section{Conclusions and Limitations}\label{sec:conclusion}
In this paper, we propose a novel pipeline that utilizes partial sketches as visual control for inpainting partially corrupted objects within a frozen text-guided Stable Diffusion model. Our pipeline integrates three key components: a Masked Image Encoder that incorporates masks and uncorrupted contexts into the denoising latent process, a Sketch-Conditional Encoder that extracts multi-scale sketch features, and a core multi-scale Sketch-guided Bidirectional Feature Interaction (SBFI) module that fuses sketch-derived features with noisy features modulated by uncorrupted contexts. This design ensures consistent sketch-based control and enhances visual-semantic consistency with uncorrupted regions during denoising. Extensive experiments on the CUB-sketch and MSCOCO-sketch datasets demonstrate the superior performance of our approach through both quantitative and qualitative results.



\section*{Acknowledgment}
This work was supported by the Chinese Scholarship Council under Grant 202006330009.

\bibliographystyle{IEEEtran}
\bibliography{IEEE}

\end{document}